\documentclass[conference]{IEEEtran}
\IEEEoverridecommandlockouts
\usepackage{cite}
\usepackage{amsmath,amssymb,amsfonts}
\usepackage{algorithmic}
\usepackage{graphicx}
\usepackage{textcomp}
\usepackage{xcolor}
\usepackage{booktabs}
\usepackage{caption}
\usepackage{subcaption}
\usepackage{multirow}
\usepackage{pifont}    % Para \ding
\usepackage{hyperref}

\newcommand{\xmark}{\ding{55}}
\def\BibTeX{{\rm B\kern-.05em{\sc i\kern-.025em b}\kern-.08em
    T\kern-.1667em\lower.7ex\hbox{E}\kern-.125emX}}
\begin{document}

\title{Evaluating the Impact of Compression Techniques on the Robustness of CNNs under Natural Corruptions\\
\thanks{This study was financially supported by the National Council for Scientific and Technological Development (CNPq) project grant 404825/2023-0.}
}
% Enfatizar as técnicas ao inveś do modelo
% Impacto de técnicas de compressão na robustez de modelos de redes convolucionais

\author{\IEEEauthorblockN{1\textsuperscript{st} Itallo Patrick Castro Alves Da Silva}
\IEEEauthorblockA{\textit{Computing Institute} \\
\textit{Federal University of Alagoas}\\
Maceió, Brazil \\
0009-0008-8543-7776}
\and
\IEEEauthorblockN{2\textsuperscript{nd} Emanuel Adler Medeiros Pereira}
\IEEEauthorblockA{\textit{Center of Technology} \\
\textit{Federal University of Rio Grande do Norte}\\
Natal, Brazil \\
0000-0002-6694-5336}
\and
\IEEEauthorblockN{3\textsuperscript{rd} Erick de Andrade Barboza}
\IEEEauthorblockA{\textit{Computing Institute} \\
\textit{Federal University of Alagoas}\\
Maceió, Brazil \\
0000-0002-0558-9120}
\and
\IEEEauthorblockN{4\textsuperscript{th} Baldoino Fonseca dos Santos Neto}
\IEEEauthorblockA{\textit{Computing Institute} \\
\textit{Federal University of Alagoas}\\
Maceió, Brazil \\
0000-0002-0730-0319}
\and
\IEEEauthorblockN{5\textsuperscript{th} Márcio de Medeiros Ribeiro}
\IEEEauthorblockA{\textit{Computing Institute} \\
\textit{Federal University of Alagoas}\\
Maceió, Brazil \\
0000-0002-4293-4261}
}

\maketitle

\begin{abstract}
Compressed deep learning models are crucial for deploying computer vision systems on resource-constrained devices. However, model compression may affect robustness, especially under natural corruption.
Therefore, it is important to consider robustness evaluation while validating computer vision systems.
This paper presents a comprehensive evaluation of compression techniques—quantization, pruning, and weight clustering—applied individually and in combination to convolutional neural networks (ResNet-50, VGG-19, and MobileNetV2). Using the CIFAR-10-C and CIFAR-100-C datasets, we analyze the trade-offs between robustness, accuracy, and compression ratio. Our results show that certain compression strategies not only preserve but can also improve robustness, particularly on networks with more complex architectures. 
Utilizing multiobjective assessment, we determine the best configurations, showing that customized technique combinations produce beneficial multi-objective results. 
This study provides insights into selecting compression methods for robust and efficient deployment of models in corrupted real-world environments.
\end{abstract}


\section{Introduction}
%Deep neural networks and machine learning techniques have been widely used in various computer vision tasks, such as object classification. However, 
Humans can adapt to changes in image structures and styles, including snow, blur, and pixelation; however, computer vision models struggle with such variations. 
Consequently, the performance of the model decreases when the input is naturally distorted, which poses challenges in practical settings where such distortions are inevitable. 
For example, autonomous vehicles must handle diverse conditions such as fog, frost, snow, sandstorms, or falling leaves to accurately read traffic signs. 
However, predicting all natural conditions is not feasible. 
Therefore, evaluating the robustness of the model is crucial to validate the reliability of computer vision and machine learning systems in safety-critical contexts \cite{hendrycks2018benchmarking}.

The robustness of models against different types of perturbation has been a studied topic in the machine learning community.
Natural corruptions, which are an important type of disturbance, are common in real scenarios and can reduce the accuracy of models, so their study has been widely carried out \cite{hendrycks2018benchmarking, croce2021robustbench, Hendrycks_2021_ICCV}.
Models are sometimes deployed on resource-limited devices, such as embedded systems and smartphones, necessitating a reduction in model size while maintaining accuracy. Techniques such as pruning (sparsity) \cite{zhu2017prune}, quantization \cite{jacob2018quantization}, and weight sharing (clustering) \cite{han2016deepcompression} have been suggested for this purpose. These methods can be used individually or in combination to take advantage of their unique strengths in reducing the size of the model \cite{tensorflow2015-whitepaper}. 

Therefore, it is important to study the robustness of these compressed models against natural corruptions, as they will be used in environments prone to corrupted images, exposing potential vulnerabilities.
Works such as \cite{Mitra_2024_CVPR, xiao2023robustmq, effect2024quantization,austria2025compoudedmodels} applied compression techniques to machine learning models and evaluated the robustness of these optimized models. 
Although \cite{austria2025compoudedmodels} applies two or more successive techniques to reduce model, their study focuses on robustness against adversarial attacks, and the combinations of techniques explored were more limited.
%size and subsequently evaluates the \emph{trade-off} between robustness and the compression level achieved, 
% However, no work has applied two or more techniques successively to the model to achieve a greater reduction in its size and, after that, evaluate the \emph{tradeoff} between robustness and the level of compression achieved by the optimized model.

The purpose of this study is to analyze the impact of model compression techniques on robustness against natural corruption. 
From there, analyze the impact of these compression techniques in relation to different models and also how robustness relates to other important metrics in relation to the compressed model. 
Our main contributions are as follows.
\begin{itemize}
    \item Evaluate the impact of compression techniques and their combinations on robustness under natural corruptions of models with different architectures.
    \item Evaluate the trade-off between robustness, accuracy, and compression ratio.
    % \item Ranking the best and worst combinations of compression techniques and models 
\end{itemize}

%todo
The structure of this paper is as follows: Section \ref{sec:related_work} surveys the literature pertinent to our research. Section \ref{sec:methodology} outlines the corruptions, compression methods, models, and evaluation criteria. Section \ref{sec:results} delivers the experimental findings and model evaluations. Finally, Section \ref{sec:conclusion} offers conclusions and future work suggestions.

\section{Related Works}
\label{sec:related_work}

In \cite{hendrycks2018benchmarking} the authors set a \textit{benchmark} for evaluating classifier robustness, culminating in the creation of IMAGENET-C. 
%and IMAGENET-P datasets. IMAGENET-C focused on robustness to corruption, while IMAGENET-P assessed resilience to perturbations. 
This \textit{benchmark} aids in gauging model performance against these common challenges. Additionally, it distinguishes between robustness against corruption and perturbation versus adversarial attacks. Other datasets like CIFAR-10-C, CIFAR-100-C, TINY IMAGENET-C, and IMAGENET 64 X 64-C serve similar purposes as IMAGENET-C. Robustness against corruption is quantitatively assessed using \textit{Mean Corruption Error} (mCE) and \textit{Relative Mean Corruption Error} (Relative mCE).

The work \cite{Mitra_2024_CVPR} evaluates the impact of \emph{post-hoc} pruning on the calibration and resilience of image classifiers facing corruptions. The research demonstrates that \emph{post-hoc} pruning in \emph{Convolutional Neural Networks} (CNN) significantly boosts calibration, performance, and durability against model corruption. Structured pruning focused on filters or channels, while unstructured pruning targeted model weights, with pruning rates spanning 10\% across 0\% to 70\%.

The work \cite{xiao2023robustmq} evaluated the robustness of quantized neural network models against various noises, such as adversarial attacks, natural corruptions, and systematic noises in ImageNet\cite{imagenet2009dataset}. The experiments showed that lower bit quantization is more resistant to adversarial attacks but becomes less resistant to natural corruptions and systematic noise.

The work \cite{effect2024quantization} analyzed the performance of quantized neural network models under various noises, including data disturbance, model parameter disturbance, and adversarial attacks, using the data set \emph{Tiny ImageNet}. The benchmark evaluates three classical architectures and the quantization method, BBPSO-Quantizer, considering low bitwidths. The authors' findings indicate that models quantized with fewer bits perform better than floating-point models.

In \cite{austria2025compoudedmodels}, the authors explore the impact of model compression methods, such as quantization, pruning, and clustering, on the adversarial robustness of neural networks, especially within TinyML. Using FGSM and PGD methods to create adversarial examples, the study assessed the robustness of models trained adversarially post-application techniques, both separately and jointly. The results indicate a general decline in robustness; however, the combination of methods does not markedly deteriorate the results beyond individual applications. Some techniques even bolstered resilience to minor perturbations.
%, hinting at potential model optimization avenues that retain or enhance robustness.
%todo
% Fazer uma tabela com os artigos citados e nas colunas as técnicas utilizadas e combinações
\begin{table}[htbp]
\centering
\caption{Comparison of papers dealing with the impact of model compression techniques on robustness under natural corruptions.}
\label{tab:related_works}
\resizebox{\columnwidth}{!}{%
\begin{tabular}{@{}ccccc@{}}
\toprule
                                                                      & Quantization              & Pruning                   & Clustering                & Combinations              \\ \midrule
\cite{Mitra_2024_CVPR}                             & \xmark     & \checkmark & \xmark     & \xmark     \\
\cite{xiao2023robustmq} & \checkmark & \xmark     & \xmark     & \xmark     \\
\cite{effect2024quantization}                        & \checkmark & \xmark     & \xmark     & \xmark     \\
\textbf{This Paper}                                                            & \checkmark & \checkmark & \checkmark & \checkmark \\\midrule
\end{tabular}
}

\end{table}

Although the work \cite{hendrycks2018benchmarking} establishes the benchmark for evaluating robustness against natural corruptions, it does not explore the impact of compression techniques on robustness. 
In addition, \cite{austria2025compoudedmodels} analyzes the impact of compression techniques on robustness, however robustness against adversarial attacks.
As summarized in Table~\ref{tab:related_works}, \cite{Mitra_2024_CVPR, xiao2023robustmq, effect2024quantization} focus on individual techniques and do not investigate combinations of methods or include weight sharing in robustness analysis.
Therefore, this work is relevant for analyzing the impact of compression techniques on the robustness of CNNs under natural corruptions, as well as for investigating how robustness interacts with other key metrics, such as accuracy and compression ratio.

\section{Methodology}
\label{sec:methodology}

Initially, models were loaded with pre-trained weights on the ImageNet dataset to meet the study objectives~\cite{imagenet2009dataset}. 
Transfer learning was then used to adapt these models to the CIFAR-10 and CIFAR-100 datasets. 
The initial evaluation with uncorrupted test images provided a baseline performance measurement. 
Various compression techniques were subsequently applied, both individually and in combination. 
The optimized models were evaluated using three key metrics: accuracy, \textit{mean corruption error} (mCE) and compression ratio. 
These metrics support a comprehensive analysis that focuses on performance, robustness, and efficiency. The following sections detail these procedures.

\begin{figure}[htbp]
    \centering
        \includegraphics[width=0.8\columnwidth]{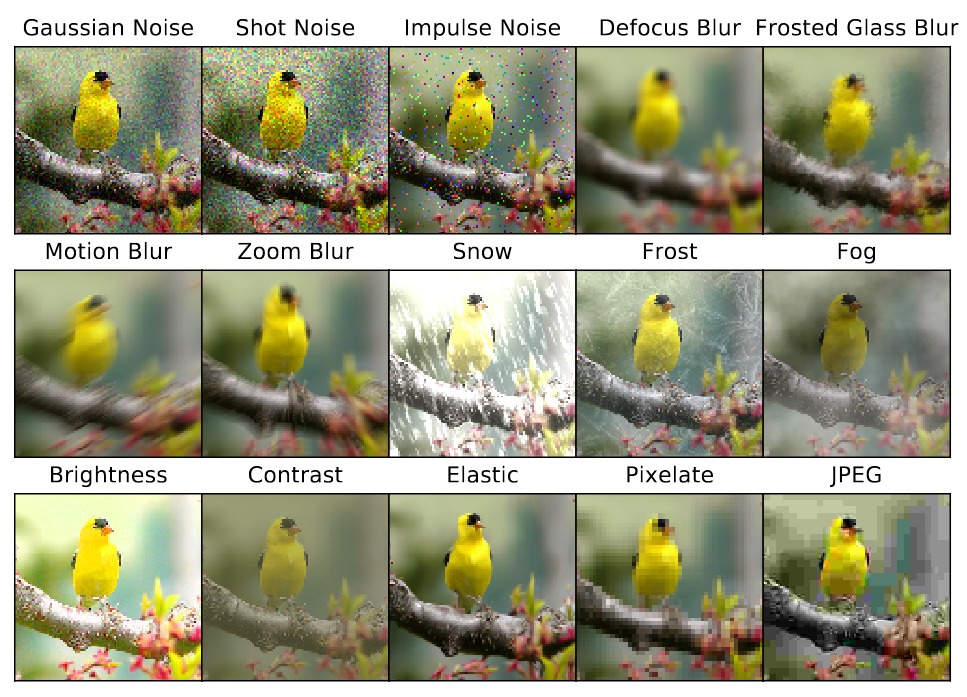}
    \caption{The 15 types of corruption considered in this work (Source: \cite{hendrycks2018benchmarking}).}
    \label{fig:corruption_types}
\end{figure}

The datasets selected for corrupted images were CIFAR-10-C and CIFAR-100-C\cite{hendrycks2018benchmarking}, publicly accessible and feature 15 types of corruption, each with 5 severity levels (see Figure \ref{fig:corruption_types}). 
Thus, each dataset includes 75 corruptions applied to CIFAR-10 and CIFAR-100\cite{krizhevsky2009learning}. 
Consequently, CIFAR-10 and CIFAR-100\cite{krizhevsky2009learning} were used to assess uncorrupted (clean) images. 
Although \cite{hendrycks2018benchmarking} proposed other image datasets, CIFAR-10-C and CIFAR-100-C were selected primarily for their smaller size compared to others. 
The CIFAR-10 and CIFAR-100 test sets each contain 10,000 images, allowing rapid model construction and validation without extensive computational resources. 
In addition, using two data sets facilitated the model evaluation in both a simpler context with CIFAR-10 (10 classification classes) and a more complex context with CIFAR-100 (100 classification classes).

To assess robustness against common corruptions, in \cite{hendrycks2018benchmarking} the authors used several state-of-the-art models, including ResNet-50 \cite{resnet_2016_CVPR} and VGG-19 \cite{vgg2014}. 
In addition, in \cite{xiao2023robustmq} the authors used MobileNetV2 \cite{sandler2018mobilenetv2}. 
In this work, we will use the following models: (1) ResNet-50, as it is a widely used classic backbone architecture in computer vision tasks and, in addition, showed the best results in \cite{hendrycks2018benchmarking}; (2) VGG-19, as it is a deep architecture with simple layers; (3) MobileNetV2, because it is a \emph{lightweight} model designed for efficient deployment in embedded devices. 
Therefore, with these options, we can have a variation in the types of models and their sizes.

\begin{figure}[htbp]
    \centering
        \includegraphics[width=\columnwidth]{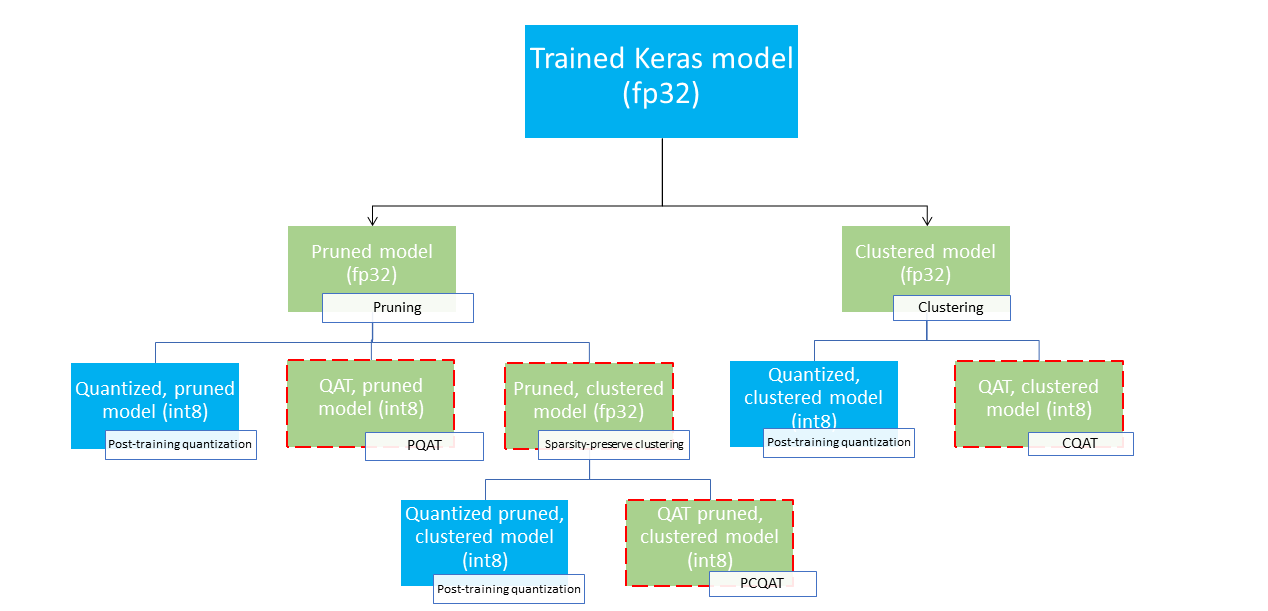}
    \caption{Tree with Tensorflow's collaborative optimizations (Source: \cite{tensorflow2015-whitepaper}).}
    \label{fig:collaborative_optimization}
\end{figure}

As tools for training, \emph{fine-tuning} and optimizing the models, we use TensorFlow \cite{tensorflow2015-whitepaper} together with LiteRT \cite{google_edge_litert}.
LiteRT is optimized for machine learning on edge devices, and TensorFlow has a dedicated module for model compression. 
Compressions are the key point of our study and TensorFlow and LiteRT have the techniques of quantization \cite{jacob2018quantization}, post-training, and during training, pruning \cite{zhu2017prune} and \emph{Weight Sharing}\cite{han2016deepcompression} with easy implementation. 
In addition, we will apply a collaborative compression by applying one technique after another to ensure the effects of the accumulated compressions.
Figure \ref{fig:collaborative_optimization} shows various deployment paths that can be explored in search of more compressed models, the leaf nodes are models ready for deployment, which means that they are partially or fully quantized in the \emph{tflite} format. 
The green fill indicates the steps where \emph{fine-tuning} is required, and the red border indicates the collaborative optimization steps, the deployment path with only quantization (post or during training) has been omitted from the figure. 
Therefore, the idea of this work is to explore both techniques individually and together to assess the impact that successive combined techniques have on the robustness against corruption. 
The set of applied techniques can be seen in Table \ref{tab:applied_techniques}.
In addition, we use the Keras API to implement the CNN models considered in this study.
Finally, we use the \textit{Python} language with the \textit{Pandas}, \textit{NumPy}, \textit{Scikit-learn}, and \textit{Matplotlib} libraries.\footnote{All the code used is available for reproduction. \href{https://github.com/itallocastro/compression-techniques-robustness-under-natural-corruptions}{https://github.com/itallocastro/compression-techniques-robustness-under-natural-corruptions}}

\begin{table}[htbp]
\centering
\caption{Set of compression techniques applied.}
\label{tab:applied_techniques}
\begin{tabular}{@{}cc@{}}
\toprule
\textbf{Code} & \textbf{Compression Technique}                                   \\ \midrule
\textbf{1}    & Original                                              \\
\textbf{2}    & Quantization Int8                                     \\
\textbf{3}    & Sparsity (Pruning)                                    \\
\textbf{4}    & Sparsity with Quantization Int8                       \\
\textbf{5}    & Clustering (Weight Sharing)                           \\
\textbf{6}    & Clustering with Quantization Int8                     \\
\textbf{7}    & Sparsity preserving Clustering                        \\
\textbf{8}    & Sparsity preserving Clustering with Quantization Int8 \\
\textbf{9}    & Quantization Aware Training (QAT)                     \\
\textbf{10}   & QAT with Quantization Int8                            \\
\textbf{11}   & Clustering preserving QAT (CQAT)                      \\
\textbf{12}   & CQAT with Quantization Int8                           \\
\textbf{13}   & Sparsity preserving QAT (PQAT)                        \\
\textbf{14}   & PQAT with Quantization Int8                           \\
\textbf{15}   & Pruning preserving Clustering preserving QAT (PCQAT)  \\
\textbf{16}   & PCQAT with Quantization Int8                          \\ \bottomrule
\end{tabular}
\end{table}

Table~\ref{tab:all_hyperparameters} presents a summary of the hyperparameters defined for each component of the model, including architecture, training configuration, and compression techniques. 
We trained the models, both original and compressed, using the hyperparameters in the \textit{Model} section of this table, we use the Adam optimizer \cite{kingma2014adam} due to its adaptive learning rate and momentum properties, which accelerate convergence and have shown strong empirical performance in training deep neural networks and the use of \emph{categorical cross-entropy} was due to the kind of problem we have. 
We used the \emph{Early Stopping} technique in all stages that involved transfer learning or \emph{fine-tuning}, with a limit of 30 epochs and patience of 5 epochs. 
However, we did not apply \emph{Early Stopping} to the \emph{pruning} technique, as it is performed progressively over the course of the epochs, and stopping training early would compromise its effectiveness \cite{zhu2017prune}. 
For the \emph{pruning} and \emph{weight sharing} techniques, we followed the standard hyperparameters defined by TFLite, as specified in the \textit{Pruning} and \textit{Clustering} sections of Table~\ref{tab:all_hyperparameters}.

\begin{table}[htbp]
\centering
\caption{Summary of hyperparameters used in the experiments.}
\label{tab:all_hyperparameters}
\resizebox{\columnwidth}{!}{%
\begin{tabular}{@{}lll@{}}
\toprule
\textbf{Category} & \textbf{Hyperparameter}          & \textbf{Value}                          \\ \midrule

\multirow{4}{*}{\textit{Layers}} 
    & Input Layer                               & (32, 32, 3)                              \\
    & Up Sampling 2D                            & \textbf{Size}: (7, 7)                    \\
    & Intermediate Dense Layer                 & \textbf{Units}: 1024; Activation: relu   \\
    & Final Dense Layer                        & \textbf{Units}: 10 or 100; Activation: \emph{Softmax} \\ \midrule

\multirow{2}{*}{\textit{Model}}  
    & Optimizer                                 & Adam                                    \\
    & Loss function                             & \emph{Categorical Cross-Entropy}        \\ \midrule

\multirow{3}{*}{\textit{Pruning}} 
    & Target sparsity                           & 50\%                                     \\
    & Start \emph{Step}                         & 0                                        \\
    & End \emph{Step}                           & Until the end of training                \\ \midrule

\multirow{2}{*}{\textit{Clustering}} 
    & Number of \emph{Clusters}                 & 16                                       \\
    & Centroid initialization                  & \emph{KMEANS++}                          \\

\bottomrule
\end{tabular}
}
\end{table}

\begin{table}[htbp]
\centering
\caption{Results original models.}
\label{tab:result_original}
\begin{tabular}{@{}lccc@{}}
\toprule
             & \textbf{CIFAR-10}                          & \textbf{CIFAR-100}                         & \multicolumn{1}{l}{}          \\ \midrule
\textbf{Model}        & \multicolumn{1}{l}{\textbf{Accuracy (\%)}} & \multicolumn{1}{l}{\textbf{Accuracy (\%)}} & \multicolumn{1}{l}{\textbf{Size (MB)}} \\ \midrule
ResNet-50    & 88\%                              & 76\%                              & 91MB                          \\
VGG-19       & 72\%                              & 56\%                              & 73MB                          \\
MobileNetV2 & 90\%                              & 72\%                              & 13MB                          \\ \bottomrule
\end{tabular}
\end{table}

Table \ref{tab:result_original} presents the accuracy and dimensions of the original models before compression. 
These results will serve as a reference for the compressed models. 
In particular, for CIFAR-10, MobileNetV2 achieves the highest accuracy at 90\% and is the smallest model, while VGG-19 performs the worst at 72\%. 
Regarding CIFAR-100, ResNet-50 leads with a 76\% accuracy and is the largest model, while VGG-19 again shows the lowest performance at 56\%.

We considered the following metrics to evaluate models and compression techniques: accuracy, \emph{Mean Corruption Error} (mCE), \cite{hendrycks2018benchmarking} and the compression ratio of the model.
The \emph{Corruption Error} (CE) is calculated using Equation \ref{eq:mCE}.

\begin{equation}
\text{CE}_c^f = \frac{\sum_{s=1}^5 E_{s,c}^f}{\sum_{s=1}^5 E_{s,c}^\text{Baseline}}
\label{eq:mCE}
\end{equation}
where $c$ indicates which of the 15 corruptions is being evaluated, $f$ is the compressed model being evaluated (e.g. ResNet-50 (Pruned)), $s$ is the severity level ranging from 1 to 5, $E_{s,c}^f$ is the model error rate for the corrupted images and $Baseline$ is the original model trained with the clean data.
From this, we can summarize the robustness of the model by calculating the average of the 15 CE (${\text{CE}}_{\text{Gaussian Noise}}^f, {\text{CE}}_{\text{Shot Noise}}^f, ..., {\text{CE}}_{\text{JPEG}}^f$). 
This result is \emph{Mean Corruption Error} (mCE).
The compression ratio is obtained from the ratio between the original model size and the compressed model size.

We will compute the Pareto front for the obtained solutions using these metrics, focusing on a multi-objective evaluation aimed at maximizing accuracy, minimizing mCE, and maximizing the compression ratio. 
The Pareto front will present a set of non-dominated solutions that surpass other solutions within the \cite{tuvsar2014visualization} search space. 
A Pareto-optimal solution cannot improve any objective without negatively impacting at least one other objective.

\section{Results and discussion}
\label{sec:results}

\begin{figure}[t]
    \centering
    \begin{subfigure}[b]{\columnwidth}
         \centering
         \includegraphics[width=\columnwidth]{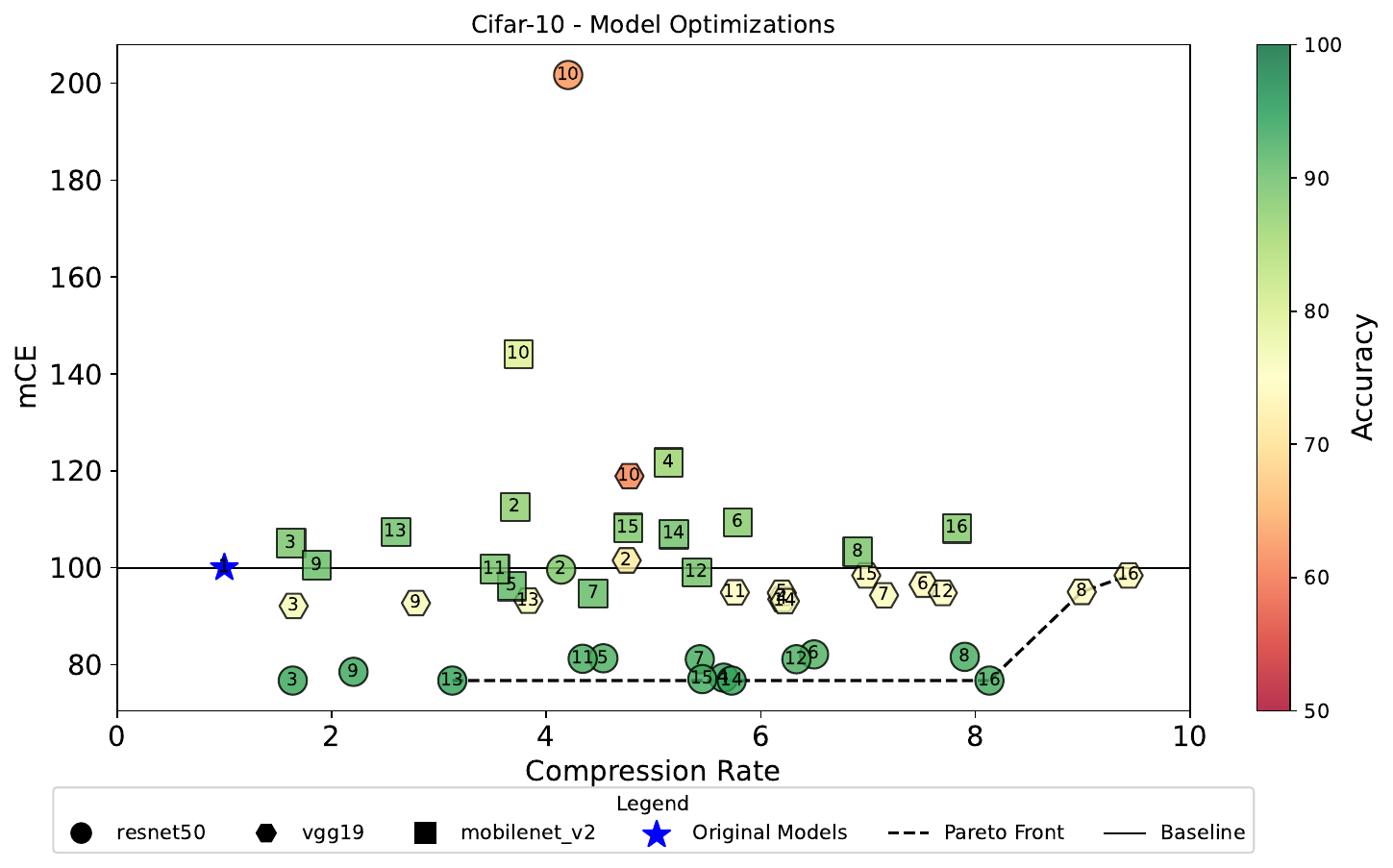}
         \caption{CIFAR-10}
         \label{fig:results_cifar10}
     \end{subfigure}
     \begin{subfigure}[b]{\columnwidth}
         \centering
         \includegraphics[width=\columnwidth]{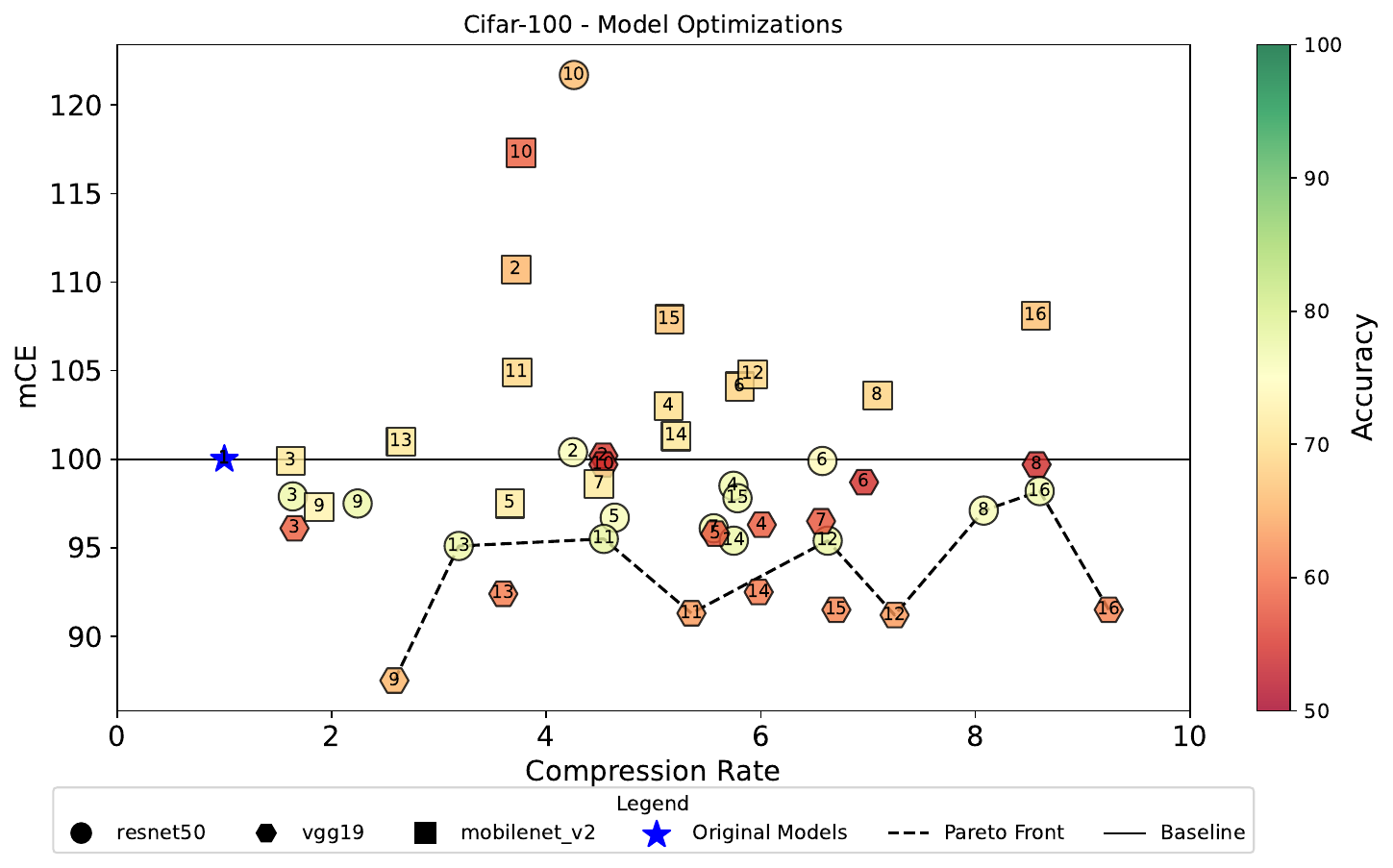}
         \caption{CIFAR-100}
         \label{fig:results_cifar100}
     \end{subfigure}
    \caption{Graphical representation of the optimization results for (a) CIFAR-10 and (b) CIFAR-100, considering the multi-objective evaluation of mCE (mean Corruption Error), compression rate, and accuracy. Each point corresponds to a specific model optimization technique, as detailed in Table~\ref{tab:applied_techniques}. The color gradient represents the accuracy, while the symbols distinguish different model architectures. The dashed line highlights the Pareto front, and the solid line denotes the baseline performance of the original models.}
    \label{fig:results_pareto}
\end{figure}

In this section, we present the results obtained in the experiments.
%The results obtained by this study can be seen in Table \ref{tab:result_original} and \ref{tab:best_worst_cifar10_100} and Figures \ref{fig:results_cifar10} and \ref{fig:results_cifar100}. 
%Table \ref{tab:result_original} shows the accuracy and size of the original models, without the application of compression techniques. 
Figure \ref{fig:results_pareto} shows the results obtained from the compressed models in terms of compression rate, mCE and accuracy, considering CIFAR-10, CIFAR-100 and the compression techniques defined in Table~\ref{tab:applied_techniques}. 
%Finally, Table \ref{tab:best_worst_cifar10_100} summarizes the best and worst results obtained for each metric evaluated, indicating the respective models and the compression techniques applied.

\begin{table*}[t]
\centering
\caption{Best and worst values for mCE, Compression Rate (CR), and Accuracy (Acc) on CIFAR-10 and CIFAR-100.}
\label{tab:best_worst_cifar10_100}

\begin{subtable}[t]{0.48\textwidth}
\centering
\caption{CIFAR-10}
\label{sub:cifar10}
\begin{tabular}{@{}cccccc@{}}
\toprule
                                & \textbf{mCE} & \textbf{CR} & \textbf{Acc} & \textbf{Model} & \textbf{Technique} \\ \midrule
\textbf{Best mCE}               & 76.7         & 8.13        & 92.67        & ResNet-50      & PCQAT Int8 (\#16)    \\
\textbf{Best CR}                & 98.4         & 9.42        & 74.19        & VGG-19         & PCQAT Int8 (\#16)    \\
\textbf{Best Acc}               & 78.3         & 5.52        & 94.34        & MobileNetV2      & PQAT Int8 (\#14)     \\
\textbf{Worst mCE}              & 201.7        & 4.2         & 62.89        & ResNet-50      & QAT Int8 (\#10)      \\
\textbf{Worst CR}               & 105.1        & 1.62        & 88.9         & MobileNetV2     & Pruning (\#3)        \\
\textbf{Worst Acc}              & 61.33        & 4.77        & 61.33        & VGG-19         & QAT Int8 (\#10)      \\ \bottomrule
\end{tabular}
\end{subtable}
\hfill
\begin{subtable}[t]{0.48\textwidth}
\centering
\caption{CIFAR-100}
\label{sub:cifar100}
\begin{tabular}{@{}cccccc@{}}
\toprule
                                & \textbf{mCE} & \textbf{CR} & \textbf{Acc} & \textbf{Model} & \textbf{Technique} \\ \midrule
\textbf{Best mCE}               & 87.5         & 2.5         & 65.4         & VGG-19         & QAT (\#9)            \\
\textbf{Best CR}                & 91.5         & 9.2         & 61.5         & VGG-19         & PCQAT Int8 (\#16)    \\
\textbf{Best Acc}               & 95.5         & 4.5         & 77.8         & ResNet-50      & CQAT (\#11)          \\
\textbf{Worst mCE}              & 130.1        & 4           & 54.7         & MobileNetv2      & QAT Int8 (\#10)      \\
\textbf{Worst CR}               & 99.9         & 1.62        & 71.3         & MobileNetv2      & Pruning (\#3)        \\
\textbf{Worst Acc}              & 99.7         & 8.5         & 54.1         & VGG-19         & Sparsity+Q Int8(\#8) \\ \bottomrule
\end{tabular}
\end{subtable}
\end{table*}

Figure \ref{fig:results_cifar10} shows that 69\% (31 of 45) of the generated models achieved an mCE below or equal to the baseline, highlighting the positive influence of compression methods on the robustness of the model with the CIFAR-10 dataset. 
In particular, only the \emph{QAT with Quantization Int8} (\#10) method resulted in mCE values exceeding 100 for the three models examined, whereas techniques such as \emph{CQAT} (\#11), \emph{CQAT with Quantization Int8} (\#12), \emph{Clustering} (\#5) and \emph{Sparsity preserving Clustering} (\#7) maintained mCE at or below the baseline. 
Similarly, Figure~\ref{fig:results_cifar100} indicates that 69\% of the solutions achieved an mCE no greater than the baseline for CIFAR-100, reaffirming the advantageous effect of compression on robustness. 
The technique \#2 yielded results above 100 for the three models concerning mCE, while techniques \#5, \#3, \#7 and \#9 consistently achieved mCE values at or below the baseline in all three models examined.

Investigation of the Pareto Front for CIFAR-10 (Figure~\ref{fig:results_cifar10}) yielded model-specific results. 
\textbf{ResNet-50} occupies 71\% of the Pareto Front (5 out of 7 solutions). The non-dominated solutions involve \textit{QAT} in conjunction with other methods, notably techniques \#13, \#14 and \#16, highlighting the combined effectiveness of QAT, Pruning and Clustering for compression, mCE, and accuracy. 
The use of \emph{Int8} quantization is a differential in this context. 
\textbf{VGG-19} makes up 29\% of the Pareto Front, with techniques \#8 and \#16, indicating that the synergy between sparsity-preserving clustering and PCQAT, bolstered by \emph{Int8} quantization, is highly effective. 
\textbf{MobileNetV2} has no representations on the Pareto Front.

Similarly, the Pareto Front for CIFAR-100 (Figure~\ref{fig:results_cifar100}) produced model-specific results.
\textbf{ResNet-50} occupies 56\% (5 out of 9 solutions) of the Pareto Front. 
The techniques that make up the Pareto Front are \#13, \#11, \#12, \#8 and \#16. 
\textbf{VGG-19} composes 44\% of the Pareto Front. 
The Pareto-optimal solutions for VGG-19 include techniques \#9, \#11, \#12 and \#16. 
The \emph{QAT} point (\#9) stands out for its very low mCE. 
%The other techniques show VGG-19's ability to optimize itself with QAT and its complex variations, achieving higher compression rates. 
The presence of \emph{PCQAT with Quantization Int8} (\#16) indicates that aggressive optimizations are possible, but the accuracy for this point in VGG-19 must be carefully evaluated. 
\textbf{MobileNetV2} did not have representation for the Pareto front. 

Table~\ref{tab:best_worst_cifar10_100} presents the extreme results that illustrate the \textit{ trade-off} dynamics among compression techniques. 
In CIFAR-10, technique \#16 stood out with the lowest mCE (76.7) for ResNet-50 and the highest compression rate (9.42) for VGG-19, demonstrating its multiobjective optimization prowess. 
Meanwhile, technique \#14 achieved the highest accuracy (94.34) on MobileNetV2, indicating varying effectiveness according to the model and the objective. 
In particular, technique \#10 yielded the worst performance in specific settings (ResNet-50 for mCE, VGG-19 for precision).
%, underscoring the importance of context in technique evaluation.
In CIFAR-100, The technique \#9 yielded the lowest mCE (87.5) on VGG-19, whereas \#10 on MobileNetV2 peaked at 130.1, highlighting the quantization effects dependent on the model and metric. 
For compression ratio, technique \#16 excelled in VGG-19 (CR of 9.2), contrasting technique \#3 on MobileNetV2, which had the lowest CR (1.62). 
Technique \#11 reached the highest accuracy (77.8) on ResNet-50. Conversely, technique \#8 on VGG-19 had the poorest accuracy (54.1). 
%This data underscores the need to align compression choices with specific objectives and models due to CIFAR-100's performance variability.

%todo
\begin{table}[htbp]
    \caption{Mean and standard deviation (in parenthesis) values for mCE, Compression Rate (CR), and Accuracy (Acc) for each compression technique and each dataset.}
    \label{tab:mean_results}
    \centering

    \begin{subtable}[t]{\columnwidth}
        \centering
        \caption{CIFAR-10}
        \label{tab:mean_cifar10}
       % \resizebox{\textwidth}{!}{%
       \begin{tabular}{@{}cccc@{}}
\toprule
 Technique & mCE          & CR         & Acc         \\ \midrule
\#1                                              & 100 (0.0)    & 1 (0.0)    & 85.2 (8.7)  \\
\#2                                     & 105.5  (6.0) & 4.1 (0.4)  & 84.2 (8.3)  \\
\#3                                               & \textbf{88.0 (13.3)}  & 1.6 (0.01) & 88.2 (8.1)  \\
\#4                        & 94.0 (19.6)  & 5.6 (0.4)  & 87.3 (8.3)  \\
\#5                                             & 89.6 (7.1)   & 4.6 (1.1)  & 87.6 (8.6)  \\
\#6                      & 94.4 (11.5)  & 6.4 (0.7)  & 86.6 (8.4)  \\
\#7                         & 88.9 (6.7)   & 5.5 (1.2)  & 87.9 (8.4)  \\
\#8  & 92.1 (9.2)   & 7.8 (0.9)  & 87.4 (8.4)  \\
\#9                                                    & 88.1 (10.3)  & 2.2 (0.4)  & \textbf{88.3 (8.3)}  \\
\#10                            & 141.8 (43.5) & 4.2 (0.4)  & 73.5 (13.9) \\
\#11                                                  & 89.9 (8.8)   & 4.4 (1.0)  & 87.3 (8.6)  \\
\#12                           & 89.7 (8.6)   & 6.3 (1.0)  & 87.4 (8.6)  \\
\#13                                                  & 88.9 (14.4)  & 3.1 (0.5)  & 88.3 (8.6)  \\
\#14                           & 88.8 (14.2)  & 5.6 (0.4)  & 88.2 (8.6)  \\
\#15                                                 & 92.2 (13.8)  & 5.5 (1.0)  & 87.1 (8.9)  \\
\#16                         & 92.1 (13.9)  & \textbf{8.3 (0.8)}  & 87.1 (8.9)  \\ \bottomrule
\end{tabular}
       % }
    \end{subtable}

    \vspace{0.5em} % espaço vertical entre as tabelas

    \begin{subtable}[t]{\columnwidth}
        \centering
        \caption{CIFAR-100}
        \label{tab:mean_cifar100}
        %\resizebox{\textwidth}{!}{%
        \begin{tabular}{@{}cccc@{}}
\toprule
                                                          & mCE                 & CR                 & Acc                 \\ \midrule
\#1                                               & 100.0 (0.0)         & 1.0 (0.0)          & 69.1 (9.0)          \\
\#2                                      & 109.0 (11.5)        & 4.1 (0.3)          & 64.4 (8.6)          \\
\#3                                               & 98.0 (1.5)          & 1.6 (0.01)         & 70.1 (8.1)          \\
\#4                        & 105.4 (12.6)        & 5.6 (0.4)          & 66.5 (8.2)          \\
\#5                                             & 96.5 (0.7)          & 4.5 (0.8)          & 69.7 (8.2)          \\
\#6                      & 103.6 (5.9)         & 6.4 (0.5)          & 66.3 (8.3)          \\
\#7                         & 97.1 (1.1)          & 5.4 (0.8)          & 69.7 (8.2)          \\
\#8  & 102.6 (5.6)         & 7.8 (0.6)          & 66.8 (9.1)          \\
\#9                                                    & \textbf{94.6 (4.8)} & 2.2 (0.3)          & \textbf{72.7 (5.1)} \\
\#10                            & 117.2 (12.8)        & 4.1 (0.3)          & 58.5 (5.5)          \\
\#11                                                  & 98.7 (6.4)          & 4.4 (0.7)          & 70.3 (6.0)          \\
\#12                           & 98.6 (6.4)          & 6.6 (0.5)          & 70.2 (6.0)          \\
\#13                                                  & 97.1 (4.1)          & 3.1 (0.4)          & 70.5 (7.0)          \\
\#14                           & 97.3 (4.0)          & 5.6 (0.3)          & 70.5 (7.1)          \\
\#15                                                 & 100.5 (7.4)         & 5.7 (0.7)          & 68.7 (6.3)          \\
\#16                         & 101.2 (7.8)         & \textbf{8.6 (0.4)} & 68.6 (6.3)          \\ \bottomrule
\end{tabular}
       % }
    \end{subtable}
\end{table}

Table \ref{tab:mean_results} presents the mean and standard deviation of the metrics evaluated for each compression technique used in the models. 
Regarding CIFAR-10 (Table \ref{tab:mean_cifar10}), technique \#3 achieved the lowest mean mCE, technique \#16 exhibited the highest mean compression rate, and technique \#9 reached the highest mean accuracy. 
In particular, 86\% (13 of 15) of the techniques showed an mCE below the original model (\#11).
%, indicating improved robustness. 
For CIFAR-100 (Table \ref{tab:mean_cifar100}), technique \#9 also yielded the lowest mean mCE and the highest mean accuracy, while technique \#16 achieved the highest mean compression rate. 
Here, 53\% (8 out of 15) of the techniques yielded mCE below the model without compression (\#1).
%, still enhancing robustness despite being less effective than with CIFAR-10.

\emph{Discussion}.
The findings indicate that the applied compression techniques generally improved model robustness against natural corruptions, as evidenced by decreased or unchanged mCE values compared to the original models in most scenarios. In particular, complex architectures like ResNet-50 and VGG-19 adopted various compression strategies more effectively. These networks exhibited substantial improvements in robustness and efficiency, particularly when advanced techniques such as PQAT, CQAT, PCQAT, and their Int8 quantized versions were used. The range of Pareto-optimal points in these models highlights the adaptability of these architectures in balancing conflicting goals. In contrast, MobileNetV2 lacks representation on the Pareto front, likely due to its compact design, limiting the advantages of these methods or requiring more specific compression approaches. It is important to mention that QAT with Quantization Int8 (\#10) did not perform well compared to other techniques, suggesting that it may not improve robustness or offer a favorable balance among precision, robustness, and compression rate. 

Therefore, these results emphasize the importance of assessing the trade-offs between robustness, accuracy, and compression rate when deploying compressed models on resource-constrained devices. For inherently small models, additional compression might be unnecessary unless severe limitations apply. In contrast, for larger models, compression could efficiently balance these criteria. In embedded systems, initiating with aggressive compression approaches tailored to the architecture can yield competitive performance within this multiobjective context.

%These results indicate the importance of careful selection of compression techniques, taking into account both the final objective (robustness, accuracy or efficiency) and the structural characteristics of the target architecture. In addition, they highlight the relevance of multi-objective analysis to guide more informed decisions in the compression process, especially in contexts where there are severe restrictions on computational resources.

\section{Conclusion and future works}
\label{sec:conclusion}

This work aimed to analyze the impact of model compression techniques on the robustness to natural corruptions. The results demonstrated that, in general, compression techniques preserve or improve robustness, which is particularly relevant for the deployment of models on resources-constrained devices. Furthermore, the findings highlighted that the effect of each technique can vary depending on the model, emphasizing the importance of tailoring compression strategies to specific architectures. Finally, when compressing models with the goal of maintaining robustness, it is crucial to consider additional metrics such as accuracy and compression rate. A joint analysis of these factors can provide deeper insight into the overall performance and trade-offs of the applied techniques.

Future work could explore the use of alternative frameworks such as \emph{PyTorch}, since the LiteRT library \cite{google_edge_litert} currently lacks support to compress some of the more recent model architectures. 
Furthermore, future studies could explore each technique in greater depth, applying new hyperparameter ranges different from those used in this study.
Moreover, a more detailed analysis of each type of corruption could be conducted to identify which distortions most significantly affect model robustness under different compression techniques. 
Evaluating performance in other corruption scenarios or using more realistic datasets may also improve the generalization of the findings. 
Finally, extending the analysis to include additional state-of-the-art models could further validate and strengthen the conclusions drawn from this study.

This study advances the scientific field by thoroughly examining the effects of model compression on robustness to natural corruption. It underscores the necessity of assessing robustness in conjunction with metrics like accuracy and compression ratio, particularly for deployment on devices with limited resources.

\bibliographystyle{ieeetr}
\bibliography{IEEEabrv,bib}

\end{document}